\documentclass{article}
\usepackage{ijcai24}
\usepackage{amssymb}

\usepackage{url}
\usepackage{array}
\usepackage[T1]{fontenc}    
\usepackage{nicefrac}       
\usepackage{microtype}      
\usepackage{xcolor}         
\usepackage{multicol}
\usepackage{multirow}
\usepackage{makecell}
\usepackage{times}
\usepackage{helvet}
\usepackage{courier}
\usepackage{soul}
\usepackage{subfigure}
\usepackage{bbding} 
\usepackage[mathscr]{euscript}
\usepackage{wrapfig}
\usepackage{babel}
\usepackage{enumitem}
\usepackage{changepage}

\usepackage{siunitx}
\usepackage[ruled,linesnumbered]{algorithm2e}

\usepackage{times}
\usepackage{soul}
\usepackage{url}
\usepackage[hidelinks]{hyperref}
\usepackage[utf8]{inputenc}
\usepackage[small]{caption}
\usepackage{graphicx}
\usepackage{amsmath}
\usepackage{amsthm}
\usepackage{booktabs}
\usepackage[switch]{lineno}

\newtheorem{theorem}{Theorem}
\newtheorem{definition}{Definition}

\title{Rethinking the Effectiveness of Graph Classification Datasets in Benchmarks for Assessing GNNs}

\author{
Zhengdao Li$^{1,2}$\and
Yong Cao$^3$\and
Kefan Shuai$^2$\and
Yiming Miao$^2$\thanks{Corresponding author}\And
Kai Hwang$^2$\\
\affiliations
$^1$Guangzhou University, Guangzhou, China\\
$^2$The Chinese University of Hong Kong, Shenzhen, China\\
$^3$Huazhong University of Science and Technology, Wuhan, China\\
}

\begin{document}

\maketitle

\begin{abstract}
Graph classification benchmarks, vital for assessing and developing graph neural networks (GNNs), have recently been scrutinized, as simple methods like MLPs have demonstrated comparable performance. This leads to an important question: Do these benchmarks effectively distinguish the advancements of GNNs over other methodologies? If so, how do we quantitatively measure this effectiveness? In response, we first propose an empirical protocol based on a fair benchmarking framework to investigate the performance discrepancy between simple methods and GNNs. We further propose a novel metric to quantify the dataset effectiveness by considering both dataset complexity and model performance. To the best of our knowledge, our work is the first to thoroughly study and provide an explicit definition for dataset effectiveness in the graph learning area. Through testing across 16 real-world datasets, we found our metric to align with existing studies and intuitive assumptions. Finally, we explore the causes behind the low effectiveness of certain datasets by investigating the correlation between intrinsic graph properties and class labels, and we developed a novel technique supporting the correlation-controllable synthetic dataset generation. Our findings shed light on the current understanding of benchmark datasets, and our new platform could fuel the future evolution of graph classification benchmarks.
\end{abstract}

\section{Introduction}
Graph Neural Networks (GNNs) have exhibited superior performance in various domains, including recommendation system \cite{wu2022graph}, molecule property prediction \cite{wieder2020compact}, and natural language processing \cite{wu2021deep}, etc. To evaluate GNN models in these tasks, specific datasets are often selected as benchmark datasets. Given this mission, a high-quality benchmark dataset should be capable to differentiate the advancements of diverse models. For example, current available benchmarks, such as OGB \cite{huO2020}, TUDataset \cite{morrisTu2020}, etc., serve for various link-wise, node-wise, and graph-wise tasks evaluation, as well as graph classification, aiming to automatically discover the optimal method for given tasks. These datasets and benchmark frameworks have immensely facilitated GNN research.

However, recent studies \cite{erricaA2019,zhao2019learning,huO2020,dwivediB2020,morrisTu2020} have shown that GNNs may not consistently surpass other baseline methods in specific graph classification tasks. Some simple baseline methods can achieve performance similar to GNNs, and sometimes even better. For example, in the benchmark of \cite{erricaA2019}, the MoleculeFingerprint baseline outperforms significantly the widely used GNN models such as GIN \cite{xu2018powerful}, GraphSage \cite{hamilton2017inductive} on three out of four molecular datasets. Nevertheless, despite current research primarily having made significant achievements in analyzing the theoretical expressive power of GNNs \cite{xu2019How,Feng2022How,wang2022powerful} and training schemes \cite{Duan2022Bench}, the reasons for GNNs' failures from these evidences have not been thoroughly analyzed. Few researchers are paying attention to the issues inherent in the datasets themselves. 

Therefore, we have adopted a different perspective: dataset compatibility. Our investigation focuses on whether the datasets themselves are suitable for evaluating the advancements of Graph Neural Networks (GNNs) compared to other methods. This aspect is critical for a fair assessment of whether a GNN method has truly shown improvement. Studies in neural language processing, like \cite{xiao-etal-2022-datasets}, define effectiveness as the performance variance across different methods. However, this definition is not directly applicable to graph classification problems. For example, in binary classification problems versus 10-class classification problems, the absolute values of variance are not directly comparable. Hence, in our paper, we reevaluate the effectiveness of existing datasets and attempt to address the following two questions:

\textbf{RQ1: Can commonly used graph classification datasets serve the benchmarking purpose which is to effectively distinguish advancements of GNNs compared with other methods?} To address this question, we propose an empirical protocol (Sec.\ref{emp_protocol}) to investigate the performance disparity between baseline methods and GNN-based methods in terms of the structures and attributes separately by restricting the information input types, i.e., structural information or attributed information. Specifically, we re-organize 16 real-world datasets (Sec.\ref{data_collection}) from common benchmarks across diverse scales and application domains, and conduct extensive experiments with the proposed protocol to investigate the performance gaps fairly on a well-developed benchmark framework\footnote{https://github.com/ICLab4DL/GNNBenchEffectiveness} extended from \cite{erricaA2019}, with our new improvements: 1) supporting datasets from other benchmarks such as OGB, TUDatasets, and synthetic datasets. 2) supporting the construction and combination of various artificial node features, not limited to the framework proposed by \cite{cuiOn2022}, helps to investigate the impact of different information inputs on the performance of GNNs.

\textbf{RQ2: How to measure the effectiveness of existing graph classification datasets?}  To answer this question, we design a novel metric (Sec.\ref{def_e}) to quantify the effectiveness of diverse classification datasets by normalizing the performance gap into a scale-free quantity, with the consideration of the prediction difficulty of datasets and diversity in the number of class labels. The fairness and efficacy of the metric are justified on 16 datasets. For further exploration of the causes of the low effectiveness of datasets, we investigate the relationships between basic graph properties and class labels, and develop a novel approach (Sec.\ref{syn_dataset}) for generating controllable synthetic datasets, which enables precise control over the degree of correlations between graph properties and class labels. This allows us to study the effectiveness in a controlled environment, providing deeper insights across varying conditions. Additionally, inspired by \cite{xiao-etal-2022-datasets}, we further develop a straightforward yet effective regression method to predict the effectiveness (Sec.\ref{regress_E}) of a given dataset.

\section{Empirical Studies of Existing Graph Classification Datasets}

This section delves into empirical studies of existing graph classification datasets, offering a clear, concise, and engaging overview. Initially, we establish a straightforward and insightful evaluation measurement, utilizing diverse datasets to gain empirical insights that align with findings from other research. However, this simple measurement has its limitations. To address these shortcomings, we will introduce a novel metric designed specifically to overcome these limitations in Section 3.

\subsection{An Empirical Protocol for Evaluating Dataset Discriminability}\label{emp_protocol}

We propose a protocol that can fairly evaluate the ability of a graph classification dataset for discriminating the advancements of graph-aware methods including GNNs and graph-kernel based approches over baselines. An effective strategy is to evaluate the performance gap between them. If the performance of graph-aware methods and simple baselines exhibit similarity, it indicates that the dataset lacks the necessary discriminatory power, thus questioning its suitability as a benchmark. The protocol encompasses three main components: (1) the baselines and GNNs for classification; (2) the evaluation framework; and (3) the performance gaps as determined by the evaluation framework. In the rest of this section, we will delve into these three key components and introduce some notations for further usage.

\paragraph{Evaluation Framework.} The framework is built upon the benchmarking framework proposed by \cite{erricaA2019}, the detailed architecture can be found in the supplementary. This framework leverages risk assessment and model selection schemes to provide a fair comparison of GNN models using a k-fold cross-validation procedure for model assessment. Each validation procedure incorporates a model selection process with varying hyperparameters. We further enhance this basic framework in the following ways:

(1) We expand the dataset splitting schemes to support additional strategies, such as the molecular scaffold splitting scheme and user-defined splitting schemes. These offer meaningful, domain-aware splits as opposed to random splits.

(2) Our framework allows for the loading of datasets from various sources, including PyTorch Geometric, Open Graph Benchmark (OGB), as well as user-defined synthetic datasets.

(3) Drawing inspiration from the studies \cite{cuiOn2022}, we have equipped our framework to accept various compositions of graph-level or node-level statistical features as model input, moving beyond the support for only single node labels or edge labels.

\paragraph{Assessment of performance gap.} GNNs are superior to other neural network structures on graph data because of their ability to capture structure information. However, performance gaps in previous works fail to distinguish the effects of structure and attribute. To solve this problem, we decouple the performance gap into the \textit{structural gap} and \textit{attributed gap}. \textit{Structural performance gap} is denoted by $\delta_{\textbf{S}}$. It measures the difference in classification accuracy between a structure-dominated baseline and the best performance achieved by structure-aware methods without any attributed information, including graph-kernel based approaches and GNNs with artificial node attributes as input features. (e.g., node degree and random noise). \textit{Attributed performance gap} is denoted by $\delta_{\textbf{A}}$. It quantifies the accuracy difference between an attribute-dominated baselines that vary across applications and GNNs that utilize real node or edge attributes as input features. Note that GNNs with real node or edge attributes inevitably involve a part of structural information. A formal performance gap is given by the Definition \ref{def1}.

\begin{definition}\label{def1}
    Given a dataset $D$, a baseline method $\mathcal{M}^{\text{Baseline}}_{\textbf{type}}$, and a graph-based method $\mathcal{M}^{\text{Graph}}_\textbf{type}$, the performance gap $\delta_{\textbf{type}}(D, R, \mathcal{M}^{\text{Graph}}, \mathcal{M}^{\text{Baseline}})$ (simply denoted by $\delta_{\textbf{type}}$) between baseline and graph-based method is defined as:
$$
\delta_{\textbf{type}}  \triangleq R(D, \mathcal{M}^{\text{Graph}}_\textbf{type}) - R(D, \mathcal{M}^{\text{Baseline}}_\textbf{type}),\ \textbf{type} \in \{\textbf{S}, \textbf{A}\},
$$
where $R(D, \mathcal{M})$ is the numerical value of a given evaluation metric such as mean classification accuracy or mean AUC-ROC (Area Under the Receiver Operating
Characteristics), obtained by model $\mathcal{M}$ on dataset $D$.
\end{definition}

\paragraph{Choices of baselines.} Following the insights from \cite{cuiOn2022,erricaA2019}, we categorize baselines into two types based on input information: \textit{structure-dominated baselines} and \textit{attribute-dominated baselines}. This classification helps in delivering a detailed analysis as node attributes and structures contribute differently to model performance across datasets. For structure-dominated baselines, we use shallow MLPs with average graph degrees as input, denoted as $\mathcal{M}^{\text{Baseline}}_\textbf{S}$. For attribute-dominated baselines in attribute-graph datasets, we encode molecules as per \cite{huO2020} and use the MoleculeFingerprint model for classification, while non-attribute graph datasets employ a combination of pooling layers and shallow MLPs. These are represented as $\mathcal{M}^{\text{Baseline}}_\textbf{A}$.

\paragraph{Choices of graph-aware methods.} In our experiments, we carefully choose the $\mathcal{M}^{\text{Graph}}_\textbf{type}$ and $\mathcal{M}^{\text{Baseline}}_\textbf{type}$. We use a diverse range of graph-based methodologies, including GNN models and graph-kernel based methods for $\mathcal{M}^{\text{Graph}}_\textbf{type}$. Specifically, we employed the Graph Isomorphism Network (GIN) for its spatial approach, which focuses on the physical layout of the graph, accentuating local structures and node-level relationships. Concurrently, the Graph Convolutional Network (GCN) was chosen for its spectral method approximation, where node features are transformed into the spectral domain using the graph Fourier transform, allowing for a global analysis of the overall graph structure and relationships. Additionally, we utilized graph-kernel based methods including Weisfeiler-Lehman graph kernel (WL-GK) \cite{shervashidze2011weisfeiler}, the Subgraph-Matching kernel (SM-GK) \cite{kriege2012subgraph}, and the Shortest-Path kernel (SP-GK) \cite{borgwardt2005shortest}. This selection of both spatial (GIN) and spectral (GCN) methods, complemented by kernel-based techniques, provides a comprehensive and balanced evaluation of local and global graph features, crucial for the depth and breadth of our study in graph-based machine learning.

The performance gaps $\delta_{\textbf{S}}$ and $\delta_{\textbf{A}}$ can indicate the discriminating ability of the dataset and provide insights into it. In particular, a narrow gap with high accuracy from both methods suggests the dataset may be too simple to offer discrimination. Conversely, low accuracy from both methods implies the dataset's information is underutilized, necessitating a more advanced approach. A large gap indicates the dataset's strong discriminative power for these two models.

\subsection{Collection of Diverse Datasets}\label{data_collection}

\paragraph{Bio\&Chem.} In the fields of biology and chemistry, the ability to predict molecular properties, such as toxicity or biological activity of proteins plays a pivotal role in drug discovery and development. Datasets such as MUTAG, D\&D \cite{yanardagDeep2015}, PROTEINS, and NCI1 furnish a wealth of information for constructing and training machine learning models in these disciplines. Similarly, in chemical research, datasets like HIV and ENZYMES are indispensable for decoding the interactions between chemical compounds and their potential impacts on living organisms. The large-scale PPA dataset facilitates an understanding of intricate protein interactions and functions, significantly contributing to advancements in personalized medicine and therapeutic approaches.

\begin{table}[tp]
\label{tab1}
\scriptsize
\centering
\begin{tabular}{c|c|c|c|c|c}
\toprule[1pt]
\textbf{Domain} & \textbf{Dataset} & \textbf{Graphs} & \textbf{Classes} & \textbf{\makecell{Average \\ nodes}} & \textbf{Features}  \\
\midrule[1pt]
\multirow{6}{*}[-4ex]{Bio\&Chem}
&$\text{BACE}^{\heartsuit}$ & 1513 & 2 & 34.09 & 9|-    \\
&$\text{Tox21}^{\heartsuit}$ & 7831 & 2 & 18.57 & 9|3   \\
&$\text{HIV}^{\heartsuit}$ & 41,127 & 2 & 25.5 & 9|3   \\
&$\text{PPA}^{\heartsuit}$ & 158,100 & 37 & 243.4 & -|7   \\
& $\text{MUTAG}^{\bigstar}$ & 188 & 2 & 17.9 & 7|- \\
&$\text{NCI1}^{\bigstar}$ & 4,110 & 2 & 29.8 & 37|-  \\
&$\text{PROTEINS}^{\bigstar}$ & 1,113 & 2 & 39.1 & 3|-   \\
&$\text{AIDS}^{\bigstar}$ & 2000 & 2 & 15.69 & 38|-    \\
&$\text{DD}^{\bigstar}$ & 41,127 & 2 & 25.5 & 9|3   \\
&$\text{ENZYMES}^{\bigstar}$ & 600 & 6 & 32.6 & 3|-  \\
\hline 
\multirow{4}{*}[.1ex]{\text{\makecell{Social \\ science}}}
&$\text{IMDB-B}^{\bigstar}$ & 1,000 & 2 & 19.77 & - \\
&$\text{IMDB-M}^{\bigstar}$ & 1,500 & 3 & 13 & -  \\
&$\text{REDDIT-B}^{\bigstar}$ & 2,000 & 2 & 429.61 & -  \\
&$\text{COLLAB}^{\bigstar}$ & 5,000 & 3 & 74.49 & -  \\
\hline 
\multirow{2}{*}[-.2ex]{CV}
&$\text{MNIST}^{\blacksquare}$ & 55,000 & 10 & 70.6 & 3|-  \\
&$\text{CIFAR10}^{\blacksquare}$ & 45,000 & 10 & 117.6 & 5|-  \\
\bottomrule[1pt]
\end{tabular}
\caption{Summary of datasets with different scales, feature types and classification numbers in our experiments.}
\end{table}

\paragraph{Social science.} In the domain of social science, datasets like IMDB-Binary (IMDB-B), IMDB-Multi (IMDB-M), REDDIT-Binary (REDDIT-B), COLLAB are used to study and understand various aspects of social interactions and behaviors.

\paragraph{Computer vision (CV).} The MNIST and CIFAR10 have been fundamental in shaping the field of computer vision, offering a wide range of images for tasks like object recognition and classification. These two datasets can verify the positional learning ability of GNNs, as the samples are transformed from images into graphs with the super-pixels and coordinates as the node features that inherently carry the positional information of each node.

\subsection{Observations of Performance Gaps on 16 Real-world Datasets}

\begin{table*}[htbp]
\centering
\small
\begin{tabular}{l|c|c|c|c|c|c|c}
\toprule
Dataset & $ \mathcal{M}^{\text{Baseline}}_\textbf{A}$ &  $\mathcal{M}^{\text{GIN}}_\textbf{A}$ &  $\mathcal{M}^{\text{GCN}}_\textbf{A}$  & $\mathcal{M}^{\text{Baseline}}_\textbf{S}$ &  $\mathcal{M}^{\text{GraphKernel}}_\textbf{S}$ &  $\mathcal{M}^{\text{GIN}}_\textbf{S}$ &  $\mathcal{M}^{\text{GCN}}_\textbf{S}$ \\
\midrule
MUTAG    & 83.7 $\pm$ 8.35  & \textbf{84.07} $\pm$ 6.26 & 70.7 $\pm$ 6.89 & 79.18 $\pm$ 9.83 & 86.23 $\pm$ 8.50  & \textbf{\underline{86.71}} $\pm$ 4.67 & 82.86 $\pm$ 10.43 \\
PROTEINS & \textbf{\underline{74.24}} $\pm$ 3.09 & 70.97 $\pm$ 3.79 & 73.28 $\pm$ 3.22 & 60.95 $\pm$ 0.79 & \textbf{72.50} $\pm$ 2.58       & 68.24 $\pm$ 4.39 & 64.29 $\pm$ 2.6   \\
HIV     & 96.58 $\pm$ 0.1  & \textbf{\underline{96.86}} $\pm$ 0.13 & 96.69 $\pm$ 0.06 & 96.49 $\pm$ 0.01 & 51.00 $\pm$ 0.00    & \textbf{96.74} $\pm$ 0.09 & 96.49 $\pm$ 0.09  \\
PPA     & 20.12 $\pm$ 0.0  & \textbf{24.05} $\pm$ 0.0  & 16.08 $\pm$ 0.0  & 9.28 $\pm$ 0.0   &  -                  & 64.19 $\pm$ 0.0  & \textbf{\underline{66.72}} $\pm$ 0.0   \\
D\&D      & \textbf{\underline{76.12}} $\pm$ 2.78 & 72.22 $\pm$ 3.18 & 70.49 $\pm$ 2.13 & 62.29 $\pm$ 2.55 & 62.39 $\pm$ 1.89    & 62.73 $\pm$ 2.23 & \textbf{64.03} $\pm$ 2.64  \\
ENZYMES & 29.67 $\pm$ 5.74 & \textbf{\underline{41.78}} $\pm$ 3.92 & 31.72 $\pm$ 4.54 & 17.56 $\pm$ 1.93 & 25.00 $\pm$ 3.33    & \textbf{28.33} $\pm$ 4.24 & 22.0 $\pm$ 3.48   \\
NCI1    & 66.76 $\pm$ 1.98 & \textbf{\underline{80.54}} $\pm$ 1.16 & 76.85 $\pm$ 2.78 & 50.58 $\pm$ 1.02 & 62.50 $\pm$ 1.79    & \textbf{75.55} $\pm$ 1.31 & 65.03 $\pm$ 2.43  \\
BACE    & 68.64 $\pm$ 4.68 & \textbf{\underline{79.95}} $\pm$ 2.9  & 73.5  $\pm$ 3.48 & 54.33 $\pm$ 0.17 & 61.84 $\pm$ 0.00    & \textbf{79.82} $\pm$ 3.18 & 61.32 $\pm$ 5.07  \\
AIDS    & \textbf{99.07} $\pm$ 0.85 & 95.68 $\pm$ 1.53 & 90.05 $\pm$ 2.27 & 89.2 $\pm$ 1.16  &  \textbf{\underline{99.55}} $\pm$ 0.52 & 95.33 $\pm$ 1.16 & 86.65 $\pm$ 2.24  \\
moltox21  & 91.05 $\pm$ 0.0  & \textbf{\underline{91.31}} $\pm$ 0.0  & 90.88 $\pm$ 0.0  & 90.43 $\pm$ 0.0  &  -                        & \textbf{90.53} $\pm$ 0.0  & 90.45 $\pm$ 0.0   \\
\hline
IMDB-B   &  NA               &  NA              & NA               & 70.63 $\pm$ 3.57 & 67.10 $\pm$ 4.76          & \textbf{70.8} $\pm$ 2.81  & 69.5 $\pm$ 2.94   \\
IMDB-M   &  NA               &  NA              & NA               & 42.31 $\pm$ 4.54 & \textbf{47.00} $\pm$ 5.84     & 45.93 $\pm$ 4.19 & 44.98 $\pm$ 4.78  \\
REDDIT-B &  NA               &  NA              & NA               & 58.33 $\pm$ 1.18 & 73.50 $\pm$ 2.05          & \textbf{89.05} $\pm$ 2.14 & 86.98 $\pm$ 2.52  \\
COLLAB   &  NA               &  NA              & NA               & 67.68 $\pm$ 0.94 & 63.92 $\pm$ 1.63          & \textbf{69.92} $\pm$ 1.09 & 69.79 $\pm$ 1.11  \\
\hline
MNIST    & 24.1 $\pm$ 0.33   & \textbf{\underline{78.48}} $\pm$ 0.72 & 54.03 $\pm$ 2.15    & 9.86 $\pm$ 0.01  & -                         & 11.74 $\pm$ 1.49 & \textbf{21.93} $\pm$ 0.35  \\
CIFAR10  & 25.27 $\pm$ 0.6   & \textbf{\underline{49.87}} $\pm$ 0.4  & 45.75 $\pm$ 0.6    & 10.0 $\pm$ 0.0   & -                         & 11.13 $\pm$ 0.99 & \textbf{13.7} $\pm$ 0.62   \\
\hline
\end{tabular}
\caption{Mean test accuracy and variations of different methods in 16 graph classification datasets.}
\label{tab_all_res}
\end{table*}

\begin{figure*}[htb]
    \centering
    \subfigure[Attributed accuracy gap]{
        \includegraphics[width=0.45\textwidth]{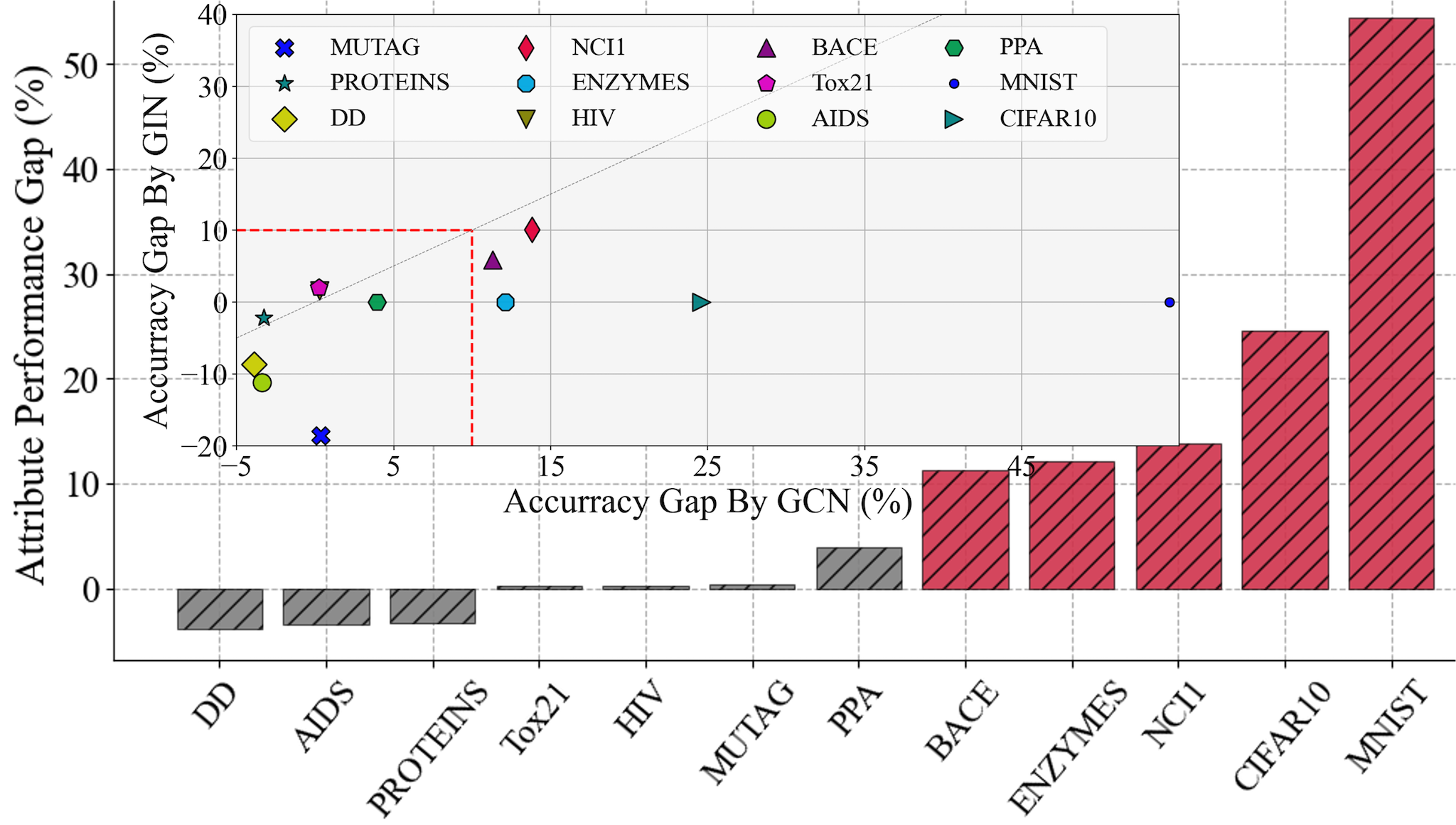}
    }
    \subfigure[Structural accuracy gap]{
        \includegraphics[width=0.51\textwidth]{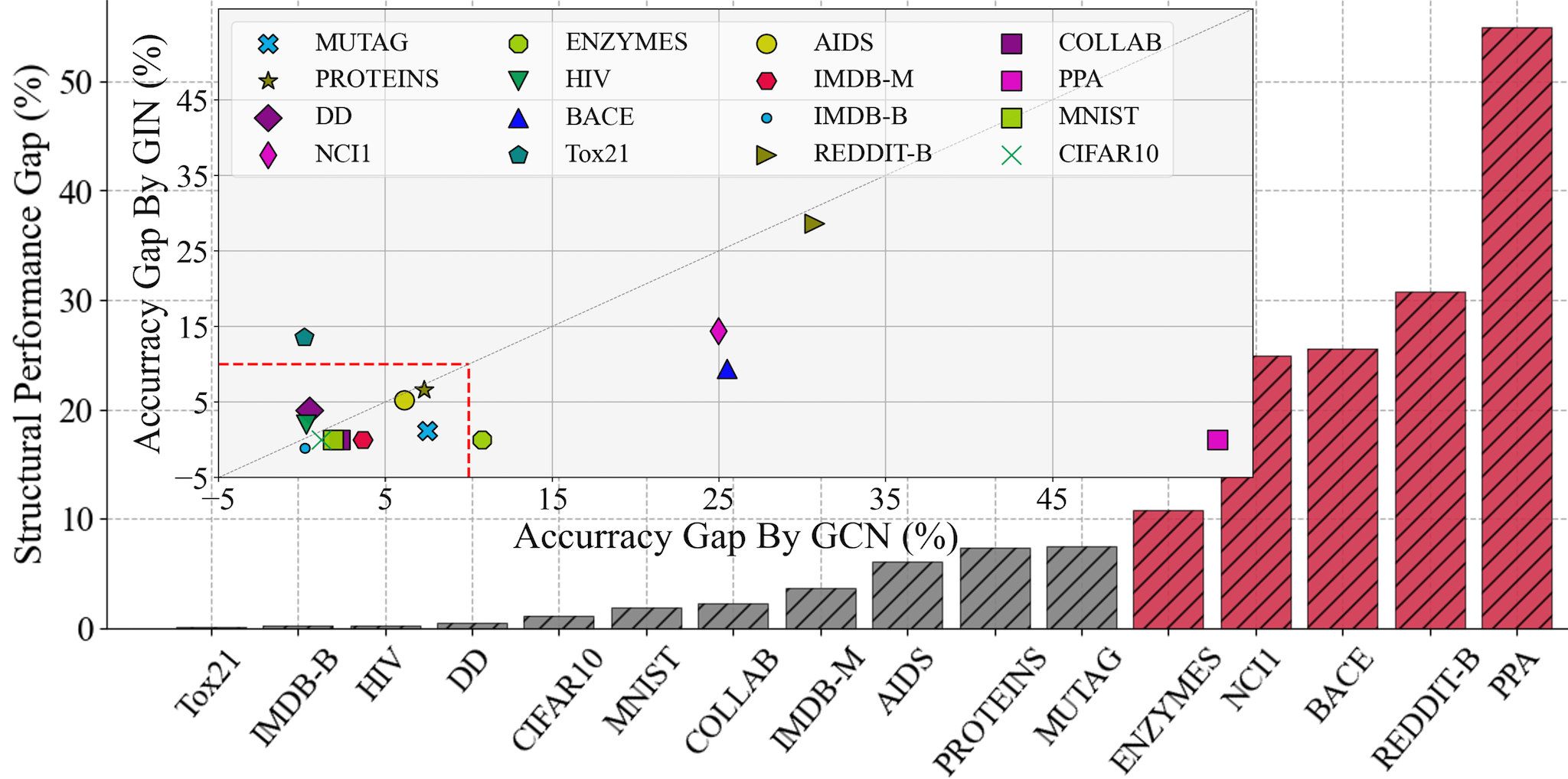}
    }
    \caption{The performance gaps on 16 graph classification datasets are categorized into two types: \textit{Ineffective} (gray) and \textit{Effective} (red) benchmarks. These are sorted in ascending order based on the size of the performance gap. An empirical threshold of 10\% is used for categorization, as observed in the inner box of each figure. This box represents the distribution of the accuracy gap for GCN and GIN.}
    \label{gap1}
\end{figure*}

\paragraph{Experimental setup.} Utilizing our proposed framework, we assessed 16 real-world datasets following the standard protocol. In Table \ref{tab_all_res}, we show the main experimental results obtained by the protocol over 16 real-world datasets, in which 14 datasets except for PPA and Tox21 were tested by 10-fold cross-validation. The baselines, GNNs and graph kernel methods are introduced in Sec\ref{emp_protocol}. (Note that, NA denotes the dataset with no attributes, - denotes the dataset is too large to run.) The values that are both bolded and underlined represent the highest accuracy across all attributed and structural models, the solely bolded values indicate the highest accuracy within one type of models. As observed from this table, GNNs excel as the state-of-the-art (SOTA) on the majority of datasets. However, it's important to highlight that the performance gap between the baseline methods and GNNs is minimal for approximately half of the datasets, which is visually represented in Figure \ref{gap1}.

In Figure \ref{gap1}(a), we depict the highest attributed accuracy gap $\delta_{\textbf{A}}$, comparing the GIN and GCN models. For molecular and protein datasets (HIV, PPA, BACE, and Tox21), we employed a baseline model formed by AtomEncoder  \cite{huO2020} and MolecularFingerprint \cite{erricaA2019}, and solely the MolecularFingerprint model for the other datasets. Subplots provide further comparisons of $\delta_{\textbf{A}}$ of GIN and GCN across different datasets. Likewise, Figure \ref{gap1}(b) reveals the greatest structural accuracy gap ($\delta_{\textbf{S}}$) among GIN, GCN, SP-GK, WL-GK, and SM-GK approaches.

From our experimental results, the following observations and insights are derived:

\noindent\textbf{Observation 1}. Most datasets excel in either attributed or structural performance gaps. Computer vision datasets MNIST and CIFAR10 showcase significant attributed performance gaps, attributable to their dependency on positional and color information of target nodes. Chemical datasets like PPA and Tox21 display noteworthy structural performance gaps due to the inadequacy of average degree information for baseline model predictions, consistent with prior findings \cite{dwivediB2020,cuiOn2022,erricaA2019,huO2020}.

\noindent\textbf{Observation 2}. Datasets displaying huge gaps for both $\delta_{\textbf{S}}$ and $\delta_{\textbf{A}}$, like ENZYMES, BACE, and NCI1, reinforce the importance of structures and specific subgraph functions in molecules and compounds. GNNs demonstrate superior performance across most of the datasets by effectively capturing both attributed and structural information simultaneously.

\noindent\textbf{Observation 3}. Interestingly, among the social science datasets, only REDDIT-B displayed a noteworthy performance gap, indicating a weak correlation between degree information and task labels. This intriguing observation will be further explored and investigated in subsequent sections.

\subsection{Limitations of Using Performance Gap as Effectiveness Measurement}

In general, half of the 16 graph classification datasets may not be effective to discriminate baselines and GNNs, by using the absolute value of performance gaps. It is important to note that solely relying on the absolute performance gap as an indicator to assess dataset effectiveness could potentially lead to some unfairness and overlook certain inherent limitations.

For instance, two binary classification datasets $D_1$ and $D_2$ have the same performance gap such as 10\%, while for $D_1$, the $R(D_1, \mathcal{M}^{\text{Baseline}})=80\%$, $R(D_1, \mathcal{M}^{\text{GNN}})=90\%$, and for $D_2$, the $R(D_2, \mathcal{M}^{\text{Baseline}})=50\%$, $R(D_2, \mathcal{M}^{\text{GNN}})=60\%$. It is obvious that the $D_2$ has more complex characteristics leading to the failures of both methods. In that case, we prefer that $D_2$ has more potential performance improvements by using advanced methods, such that it has larger effectiveness. Another limitation is the lack of consideration of the number of class labels. Suppose that $D_1$ has 2 labels, and the $D_2$ has 10 labels, the complexity of datasets is different even when the $ R(D_1, \mathcal{M}^{\text{Baseline}})=R(D_2, \mathcal{M}^{\text{Baseline}})=80\%$, and $ R(D_1, \mathcal{M}^{\text{GNN}})=R(D_2, \mathcal{M}^{\text{GNN}})=90\%$. 

Therefore, in the following section, we will deliberate on the determination of a dataset's suitability for benchmarking purposes and introduce a novel, unified metric designed to measure this degree. This metric takes into account not only the inherent complexity of the dataset and number of class labels, but also the absolute performance gaps observed among different approaches.

\section{Quantifying the Effectiveness of Benchmark Datasets}

In this section, we introduce our proposed metric designed to address the limitations of the performance gap (Definition \ref{def1}). This metric quantifies the degree to discriminate the ability of different methods. We then demonstrate the properties of this new metric and validate it on 16 benchmarks, providing answers to research questions RQ1 and RQ2.

\subsection{Dataset Effectiveness}\label{def_e}

The new metric is defined as follows and is simply denoted by $\mathcal{E}(D)$. We refer to it as Effectiveness over a dataset $D$.

\begin{definition}\label{def2}
Given a graph classification dataset $D$ which has $|Y|$ classes, and the performance gap $\delta_{\textbf{type}}(D)$ between two methods $\mathcal{M}^1$ and $\mathcal{M}^2$, the $\mathcal{E}$ to quantify the discriminating degree of $\mathcal{M}^1$ and $\mathcal{M}^2$ is defined as follows:
\begin{equation}
\begin{aligned}
    \mathcal{E}(D) =  \sum_{\textbf{type} \in \{\textbf{S}, \textbf{A}\} } \frac{|\delta_{\textbf{type}}(D)|}{R^*(|Y|-1)} \cdot \frac{1 - R^*}{1-|Y|^{-1}},
\end{aligned}
\label{eq_E}
\end{equation}
where $R^*=\min(R1, R2)$, which is the minimal value of two accuracy values from $\mathcal{M}^1$ and $\mathcal{M}^2$, denoted by $R1$ and $R2$ respectively.
\end{definition}

The Definition \ref{def2} aggregates two types of effectiveness, each type of effectiveness is the product of two components. The first component $\frac{|\delta_{\textbf{type}}(D)|}{R^*(|Y|-1)}$ is the absolute changing proportion of the performance gap which is normalized by the total number of class labels $|Y|-1$. This component varies from 0 to 1, if the worst performance is not less than random guessing.

The second component $\frac{1-R^*}{1-|Y|^{-1}}$, termed the complexity factor and denoted as $\lambda$, ranges between 0 and 1, indicating a dataset's relative complexity. The denominator, $|Y|^{-1}$, reflects random guessing accuracy, with $|Y|$ being the total task labels. The numerator represents the gap between the worst method and perfect classification. If the worst method's accuracy is near $|Y|^{-1}$, $\lambda$ nears 1, indicating high complexity. If it's near $100\%$, $\lambda$ is close to 0, suggesting a trivial dataset.

Note that, for binary datasets, $R$ can be AUC-RUC or accuracy. This is because the AUC-ROC value for random guessing is 0.5, aligning with $1-|Y|^{-1}$ when $|Y|=2$.

\begin{figure}[tp]
    \centering
    \includegraphics[width=1.0\columnwidth]{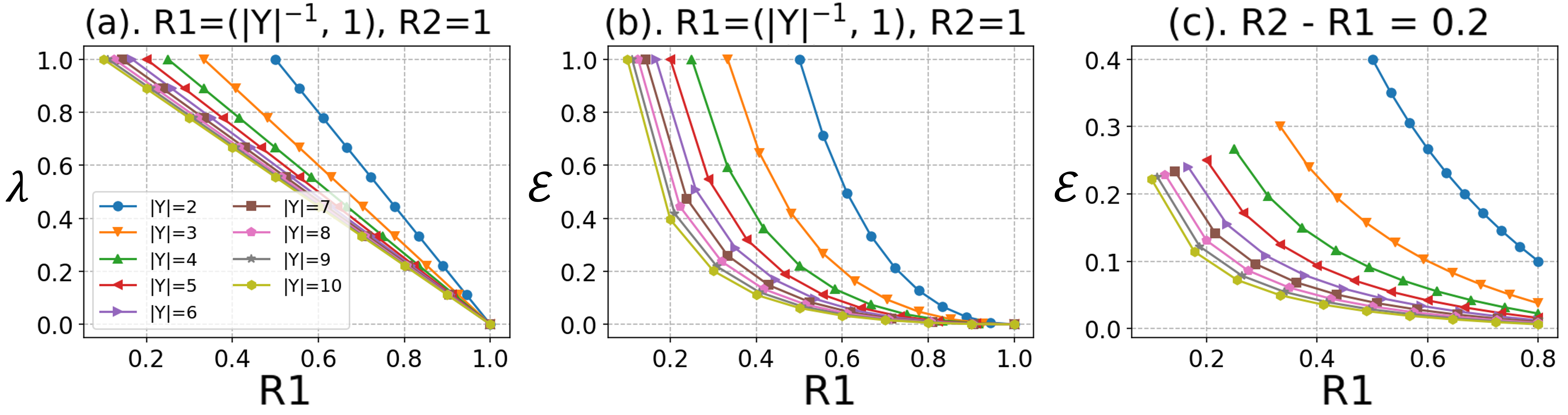}
    \caption{Properties illustration of $\lambda$ and $\mathcal{E}$.}
    \label{factor}
\end{figure}

\begin{figure}[tb]
    \centering
    \includegraphics[width=1.0\columnwidth]{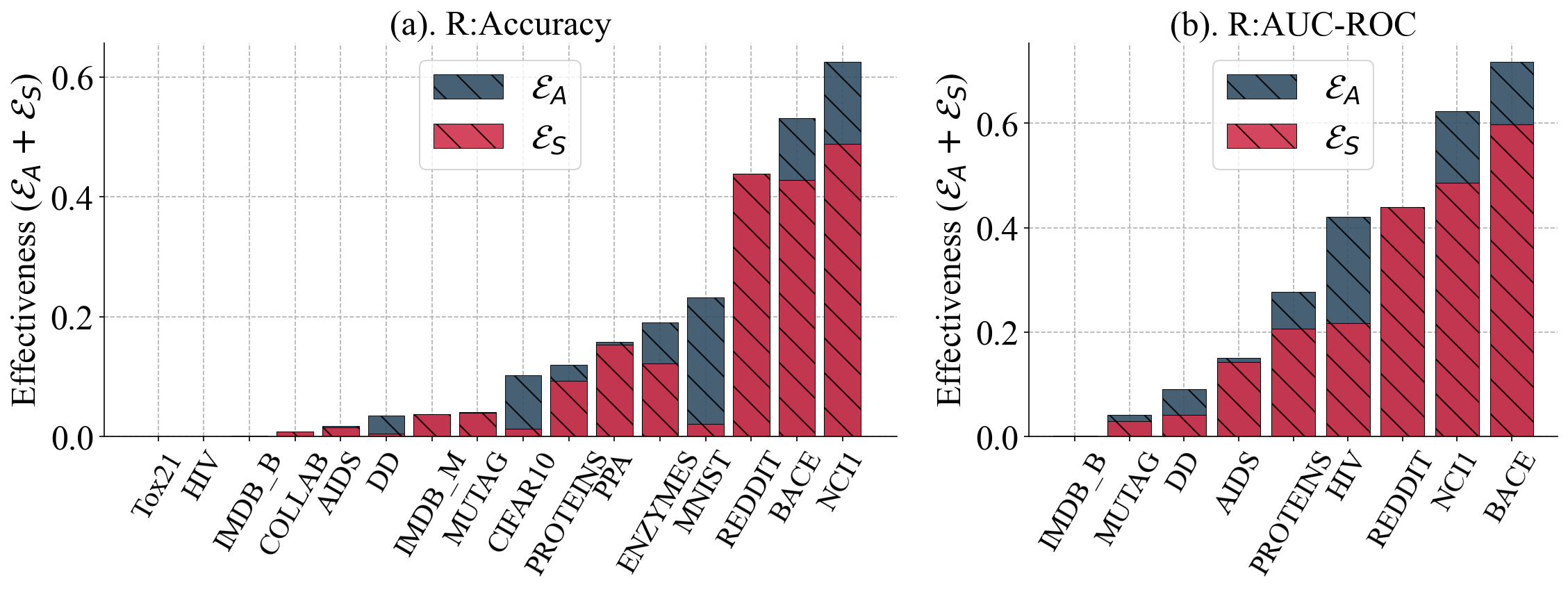}
    \caption{Effectiveness using Accuracy metric and AUC-ROC metric in terms of structural type and attributed type.}
    \label{eff}
\end{figure}

\subsection{Properties of Complexity Factor and Effectiveness}

In this section, we delve into the properties of the complexity factor $\lambda$ and how it manages dataset intricacy considering task label counts. Figure \ref{factor} elucidates properties of $\lambda$ and effectiveness $\mathcal{E}$ via variations in Eq. \ref{eq_E}.

\textbf{Property 1:} As the worst method accuracy rises, $\lambda$ linearly decreases (Figure \ref{factor}(a)). Each curve represents a dataset with task labels from 2 to 10. Essentially, a higher worst method accuracy means a simpler dataset.

\textbf{Property 2:} A smaller performance gap leads to a reduced $\mathcal{E}$ (Figure \ref{factor}(b)). As the gap decreases, the dataset's distinguishability diminishes.

\textbf{Property 3:} With a constant performance gap, $\mathcal{E}$ varies based on the worst performance values of $R1$ and $R2$ (Figure \ref{factor}(c)). Higher accuracies yield a lower $\mathcal{E}$ than lower accuracies. For instance, a 20\% difference in accuracy between two methods results in a higher $\mathcal{E}$ if the accuracies are lower.

\textbf{Property 4:} For datasets $D_1$ and $D_2$ with the same performance gaps and accuracy, if $|Y_1| < |Y_2|$, then $\mathcal{E}(D_1) > \mathcal{E}(D_2)$ (Figure \ref{factor}(a-c)). A dataset with more classes has a larger $\mathcal{E}$.

\begin{figure*}[ht]
    \centering
    \includegraphics[width=1\textwidth]{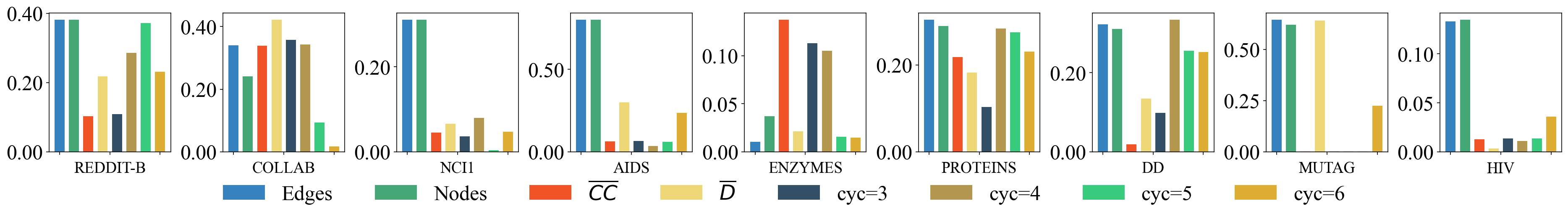}
    \caption{Correlations between graph property sequences and class labels on 9 real-world datasets.}
    \label{corr_real}
\end{figure*}

\subsection{Effectiveness of Real-world Datasets}


We examined the effectiveness $\mathcal{E}$ of 16 real-world datasets using our protocol. Figure \ref{eff}(a) shows the attributed effectiveness $\mathcal{E}\textbf{A}$ (in grey) and structural effectiveness $\mathcal{E}\textbf{S}$ (in red) for all datasets. In Figure \ref{eff}(b), we assess $\mathcal{E}$ for binary datasets using the AUC-ROC metric. While $\mathcal{E}$ values are consistent across metrics for most datasets, HIV's $\mathcal{E}$ jumps from near 0 to 0.4 with AUC-ROC, emphasizing its suitability for evaluation. Generally, $\mathcal{E}$ remains stable across different metrics. The ranking by effectiveness aligns with the performance gap, confirming a high Spearman correlation between $\mathcal{E}$ sequences and their performance gap sequences.

In conclusion, by leveraging the definition of $\mathcal{E}$ across various metrics $R$ and models $\mathcal{M}$, we can gain valuable insights. These insights aid in the assessment of a dataset's fairness and suitability for benchmarking purposes. Furthermore, this definition guides the selection of appropriate metrics and models for a given dataset, such as opting for accuracy or AUC-ROC as a metric.

\section{Investigation of Causes of Low Effectiveness of Datasets}

\subsection{Correlation Between Graph Properties and Class Labels}

Inspired by \cite{cuiOn2022,erricaA2019,huO2020}, we hypothesize that for some simple graph properties, they are highly correlated with class labels. This high correlation is what allows simple methods to achieve good accuracy. Therefore, we first examine the correlation between certain simple graph properties and class labels.

\paragraph{Graph property sequence.} We generate graph property sequences in terms of some basic graph properties such as number of nodes, average degree, count of cycles, etc. Suppose we have a non-attribute dataset $\mathbb{D}$ with $N$ samples, i.e., $\mathbb{D}=\{g_i\}_{i=1}^N$, and the corresponding labels $\mathbb{Y}=\{y_i\}_{i=1}^n$, where $y_i \in \{0, 1\}$ for a binary classification dataset. Following this sample sequence, we can generate various corresponding graph property sequences. For instance, the average degree sequence, i.e., $\overline{D}=\{\overline{d}_i\}_{n=1}^n$, where $\overline{d}_i$ is the average degree of the graph sample $g_i$. Similarly, we construct the average clustering coefficient (CC) sequence, i.e., $\overline{CC}=\{\overline{cc}_i\}_{i=1}^n$, where $\overline{cc}_i$ is the average clustering coefficient of $g_i$. Besides these two basic properties, we obtain sequences of other different graph properties, i.e., edge count sequence (denoted by Edges), node count sequence (denoted by Nodes), cycle count sequence (denoted by cyc=k), where k represents the cycle length, $k \in \{3,4,5,6\}$.

\paragraph{Correlation analysis between graph property sequences and label series.} Figure \ref{corr_real} shows the correlations between 8 graph properties and labels $\mathbb{Y}$. The correlation of Edges and Nodes with $\mathbb{Y}$ exceeds 0.2 in most datasets, often above 0.4. In molecular datasets like MUTAG, cycle count is highly correlated with labels, indicating the impact of cyclic structures. Studies \cite{chen2020can,rieck2019persistent,bouritsas2022improving} suggest WL kernels and GNNs struggle to capture substructures, underlining the importance of analyzing graph properties for method performance.

\subsection{Controllable Synthetic Datasets}\label{syn_dataset}

Real datasets are finite and insufficiently diverse for an exhaustive exploration of the effects of varying correlations between different graph properties and labels on effectiveness. Existing synthetic datasets \cite{murphy2019relational,tsitsulin2022synthetic,chen2020can}, present limitations as they rigidly utilize specific properties as labels, unable to adjust the correlation between properties and labels.

We introduce a method to generate controllable datasets, enabling precise modulation of the correlations between any graph properties and class labels. First, we propose a technique to generate random variables with a given correlation coefficient.

\paragraph{Generate correlated random variables with given coefficients.} Suppose each graph property $\mathcal{P}$, and the class label $\mathcal{Y}$ are random variables, the goal is to sample a graph property sequence $\mathbb{P}$ (e.g., $\overline{CC}$) and the class label sequence $\mathbb{Y}$ from the distributions of $\mathcal{P}$ and $\mathcal{Y}$ respectively, which satisfy a given Pearson correlation coefficient $r$ between the property and label, i.e., $r = \text{Pearson}(\mathbb{P}, \mathbb{Y})$.

\begin{theorem}
\textit{Given a set of property variables $\{\mathcal{P}_i\}_{i=1}^K$, each $\mathcal{P}_i$ follows a Gaussian distribution $\mathcal{N}(\mu_k, \sigma_k)$ or Uniform distribution $\mathcal{U}(a_k, b_k)$, and given corresponding Pearson correlation coefficients $\{r_i\}_{i=1}^{K}$ with label variable $\mathcal{Y}$, with the constraint $\sum_{i=1}^Kr_i^2 \leq 1$}, then we have:
\begin{equation}\label{eq_Y}
    \mathcal{Y} = \sigma_{\mathcal{Y}}\left(\sum_{i=1}^{K}n_ir_i + n_{0}\sqrt{1-\sum_{i=1}^{K}r_i^2} \right),
\end{equation}
\textit{where $\sigma_{\mathcal{Y}}$ is any desired standard deviation, and each $n_i$ is mutually independent and follows the same distribution as the corresponding $\mathcal{P}_i$ with the same mean value $\mu_i$ but with standard deviation equals to 1.} (The proof is based on Cholesky decomposition of a given covariance matrix.)
\end{theorem}

\begin{figure}[htp]
    \centering
    \includegraphics[width=1\columnwidth]{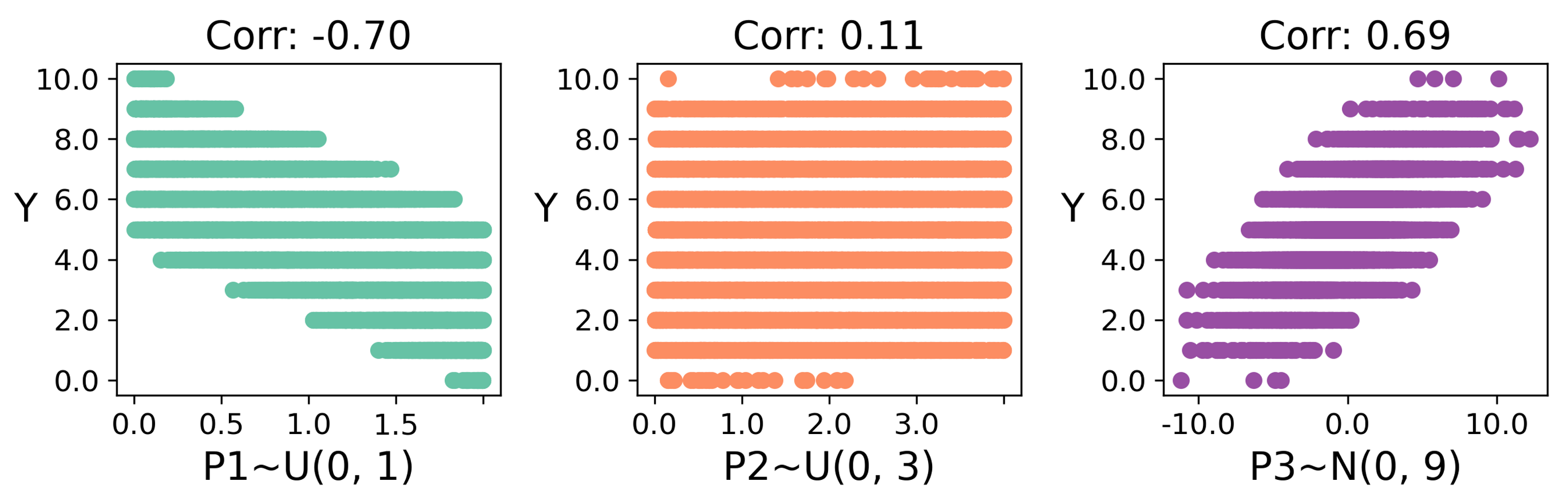}
    \caption{Generated $\mathbb{Y}$ with 11 classes by $\mathcal{P}_2, \mathcal{P}_2, \mathcal{P}_3$ following two uniform and one Gaussian distributions with the correlations ${r_1=-0.7, r_2=0.1, r_3=0.7}$ respectively.}
    \label{corr}
\end{figure}

\paragraph{Inverse graph generation by correlation.} By Theorem 1, we can easily generate a desired dataset (includes some graph properties with specific correlations with class labels) following the Algorithm \ref{alg: 1}. In Alg. 1, it is easy to prove that $\mathbb{P}_k \sim \mathcal{N}(\mu_k, \sigma_k)$, or $\mathbb{P}_k \sim \mathcal{U}(a_k, b_k)$. The \textbf{NORM} function is to normalize the $\mathcal{Y}$ into $0$ to $1$ by min-max normalization, and the \textbf{ROUND} function is to convert $\mathcal{Y}$ from decimal to an integer between 0 and $C-1$, to be used as a class label.

\begin{algorithm}[tp]
\footnotesize
\SetAlgoLined
\textbf{Input}: $\{r_k\}_{k=1}^{K}$, number of labels $C$, $\{\mathcal{P}_k\}_{k=1}^K \sim \mathcal{N}(\mu_k, \sigma_k)$ \ or \ $\mathcal{U}(a_k, b_k)$\;
\textbf{Output}:\ \text{Dataset} $\mathbb{D}$ \text{with size} $N$\;

\For{$k = 1$ \KwTo $K$}{
Sample $n_k \sim \mathcal{N}(0, \sigma_k)$\ or \ $\mathcal{U}(-\sqrt{3}, \sqrt{3})$;\\
$\mathbb{P}_k \leftarrow \mu_k+\sigma_kn_k$ or $\frac{a_k+b_k}{2}+\sqrt{\frac{b_k-a_k}{12}}n_k$; \\
}
Calculate $\mathcal{Y}$ by the Eq. \ref{eq_Y} \;
$\mathbb{Y} \leftarrow \text{ROUND}(\text{NORM}(\mathcal{Y})*C)$; \\
$\mathbb{D} \leftarrow \{(g_i, y_i)\}_{i=1}^N$, \\
where each graph $g_i$ has properties $\{\mathbb{P}_k[i]\}_{k=1}^K$, and corresponding label $y_i = \mathbb{Y}[i]$; 

\caption{Controllable dataset construction}
\label{alg: 1}
\end{algorithm}

The Figure \ref{corr} show the precise correlated relationships of generated $\mathbb{Y}$ and each properties $\mathcal{P}_1$,$\mathcal{P}_2$,and $\mathcal{P}_3$ with different correlation coefficients $r_1=-0.7, r_2=0.1, r_3=0.7$ respectively. We demonstrate the different distributions of each property. The properties follow three uniform distributions as shown in the left three boxes, and follow three normal distributions as shown in the right three boxes.

\paragraph{Construction of two synthetic datasets.} Utilizing Theorem 1, we construct two types of binary classification datasets, specifically \textbf{Syn-Degree} and \textbf{Syn-CC}. These are generated using Erdos–Renyi (ER) graphs, with a focus on controlling the average graph degree property $\overline{D}$ and average clustering coefficient property $\overline{CC}$, respectively. It's important to note that Theorem 1 is versatile and can be adapted to various graph generation processes beyond ER graphs, by defining specific numerical graph properties. We have created 9 datasets for each type, with each dataset comprising $4096$ graphs. In Syn-Degree, both $r_i^{\overline{D}}$ and $r_i^{\overline{CC}}$ range from $0.1$ to $0.9$. Conversely, in Syn-CC, all $r_i^{\overline{D}}$ are set to $0$, while $r_i^{\overline{CC}}$ varies from $0.1$ to $0.9$. Further details on the construction of these synthetic datasets are available in the supplementary materials, owing to page constraints.

\begin{figure}[h]
    \centering
    \includegraphics[width=1.0\columnwidth]{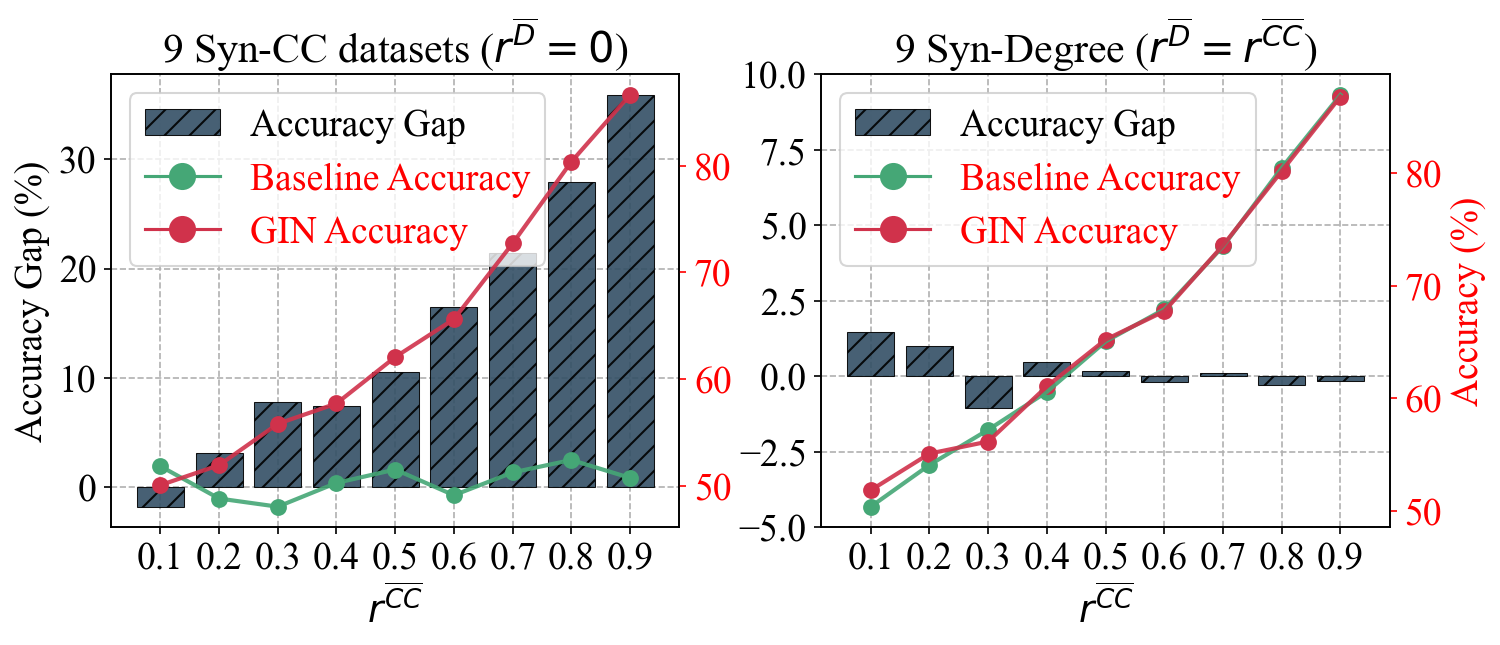}
    \caption{Controllable performance gaps by two types of synthetic datasets.}
    \label{fig_syn_gap}
\end{figure}

Under our framework, the two dataset types showed notable differences in Figure \ref{fig_syn_gap}. As correlation rises, the accuracy gap and GIN's accuracy both increase linearly, with the baseline mirroring random guessing. For the Syn-Degree dataset, GIN's accuracy and the baseline both rise linearly, keeping a minimal gap. This suggests two things: a model's prediction accuracy strongly correlates with the coefficient if it captures a graph attribute linked to the label, and GIN effectively captures clustering coefficient and degree information.

\subsection{Effective Prediction of Effectiveness Through Graph Properties and Statistical Features}\label{regress_E}

\begin{table}[tp]
\scriptsize
\centering
\begin{tabular}{cccccc}
\hline
\multirow{2}{*}{Regressor} & \multicolumn{2}{c}{Real-world datasets} & \multicolumn{2}{c}{Synthetic-CC datasets} \\ 
\cline{2-3} \cline{4-5} 
& Pearson & P-Value & Pearson & P-Value \\ 
\hline
Ridge & $0.80 \pm 0.09$ & $\le \num{1e-6}$          &  $0.87\pm 0.03$   & $\le \num{1e-6}$ \\
SVR & $0.80 \pm 0.09$ & $\le \num{1e-6}$ &  $ 0.89 \pm 0.04$ & $\le \num{1e-6}$ \\
RF & $0.89 \pm 0.03$ & $\le \num{1e-6}$   &  $0.87 \pm 0.06$  & $\le \num{1e-6}$ \\
\hline
\end{tabular}
\caption{Summary of regression results}
\label{tab_regress}
\end{table}

Most datasets show strong correlations between graph properties and labels, prompting us to explore predicting dataset effectiveness using these properties, which is computationally cheaper than benchmarking. Drawing from \cite{xiao-etal-2022-datasets}, we split each dataset into 10 distinct sets, define 26 features for graph classification, and regress effectiveness using regressors like Random Forest, SVR, and Ridge regression. Using 16 real-world datasets and 9 Syn-CC datasets, we allocate 70\% of the splits for training and 30\% for testing. Regression performance, verified by the Spearman rank coefficient in Table \ref{tab_regress}, is based on 10 repeated experiments. Both real-world and Syn-CC datasets show that basic graph properties can effectively predict dataset effectiveness.

\section{Conclusions}

Our work provides a detailed analysis of graph classification benchmarks essential for the evaluation and enhancement of GNN models. We introduced an empirical protocol to compare the performance of methods like MLPs to GNNs on certain datasets. Our novel Effectiveness metric serves as a pivotal tool for dataset validation in benchmarking. By devising a method to generate synthetic datasets, we can precisely control the correlation between graph properties and task labels, addressing the issue of low effectiveness in some benchmarks. Our efforts play a significant role in the selection of impactful benchmarks, paving the way for the development of robust GNN models and further advancements in graph learning research.

\section*{Acknowledgements}
This work was supported by National Natural Science Foundation of China (NSFC) under Grant No. 62202410, No. 62311530344, and Shenzhen Science and Technology Program under Grant JCYJ20220530143808019.
\newpage
\bibliographystyle{named}
\bibliography{ijcai24}

\end{document}